\documentclass[lettersize,journal]{IEEEtran}
\usepackage{amsmath,amsfonts,amsthm}
\usepackage{algorithmic}
\usepackage{algorithm}
\usepackage{array}
\usepackage[caption=false,font=normalsize,labelfont=sf,textfont=sf]{subfig}
\usepackage{textcomp}
\usepackage{stfloats}
\usepackage{url}
\usepackage{verbatim}
\usepackage{graphicx}
\usepackage{cite}
\newtheorem*{lemma}{Lemma:}
\newcolumntype{M}[1]{>{\centering\arraybackslash}m{#1}}

\begin{document}
\title{Pupil Learning Mechanism}

\author{Rua-Huan Tsaih,~\IEEEmembership{Member,~IEEE,} Yu-Hang Chien,~\IEEEmembership{Member,~IEEE,} and Shih-Yi Chien,~\IEEEmembership{Member,~IEEE}}
        % <-this % stops a space

\markboth{IEEE Transactions on Neural Networks and Learning Systems}%
{Rua-Huan Tsaih, Yu-Hang Chien, Shih-Yi Chien \MakeLowercase{\textit{et al.}}: Pupil Learning Mechanism}


\maketitle

\begin{abstract}
Studies on artificial neural networks rarely address both vanishing gradients and overfitting issues. In this study, we follow the pupil learning procedure, which has the features of interpreting, picking, understanding, cramming, and organizing, to derive the pupil learning mechanism (PLM) by which to modify the network structure and weights of 2-layer neural networks (2LNNs). The PLM consists of modules for sequential learning, adaptive learning, perfect learning, and less-overfitted learning. Based upon a copper price forecasting dataset, we conduct an experiment to validate the PLM module design modules, and an experiment to evaluate the performance of PLM. The empirical results indeed approve the PLM module design and show the superiority of the proposed PLM model over the linear regression model and the conventional backpropagation-based 2LNN model. 
\end{abstract}


\section{Introduction}
\IEEEPARstart{S}{tudies} of 2-layer neural networks (2LNNs) \cite{rumelhart2013learning} rarely attempt to address both vanishing gradients and overfitting issues. In this study, we follow the pupil learning procedure shown in Table \ref{plp} to derive the pupil learning mechanism (PLM), which consists of modules for sequential learning, adaptive learning, perfect learning, and less-overfitted learning.

Many learning-based artificial intelligence (AI) studies deploy artificial neural networks (ANN) with a learning algorithm to a model, but encounter difficulties. For instance, backpropagation learning \cite{rumelhart2013learning}, which naively implements gradient descent optimization (GDO) to pursue weight estimates to minimize the loss function, encounters the difficulty of vanishing gradients. Furthermore, to progressively learn in a dynamic environment under a learning-based AI paradigm, the intelligence system should be self-adaptable, i.e., able to modify its own network structure and weights without human input. Sequential learning algorithms feature online processing, adjustable parameters, and an adaptable network structure \cite{perez2018review}. Considerable efforts have been devoted to developing approaches for network structure adjustment, including (1) constructing approach: starting with a small network and then adding hidden nodes and connections \cite{ma2003new}, (2) pruning approach: starting with an extensive network and then removing irrelevant hidden nodes and connections \cite{mezard1989learning, kusuma1992cascade, chen1994generating}, and (3) combining these constructing and pruning approaches \cite{frean1990upstart, tsaih1993softening, tsaih1998explanation, tsaih2009resistant, tsai2019cramming}. However, these approaches fail to resolve overfitting challenges. An overfitted model is a model that (1) contains more parameters than can be justified by the data or (2) corresponds too closely or exactly with the training dataset and may therefore fail to fit testing data or predict future observations reliably \cite{dietterich1995overfitting, hawkins2004problem}.

\begin{table}
\renewcommand{\arraystretch}{1.3}
\begin{center}
\caption{The Pupil Learning Procedure that has the features of interpreting, picking, understanding, cramming, organizing (IPUCO)}
\label{plp}
\begin{tabular}{p{8.2cm}}
\hline
    \begin{itemize}
        \item { 
                When learning, the pupil quickly interprets instances, separating them into two groups: acquainted and unacquainted instances. Furthermore, the pupil typically learns unacquainted instances one by one and chooses easy (unacquainted) instances to learn first.
            }
        \item {
                When encountering a new (unacquainted) instance, the pupil tries to understand it by using the learning process accompanied with his current knowledge. If it cannot be understood successfully, then the pupil crams it. The cramming process adds a strict rule to his knowledge system.
            }
        \item {
                From time to time, the pupil comprehends all learned instances for a concise knowledge system.
            }
    \end{itemize}\\
\hline
\end{tabular}
\end{center}
\end{table}

Recently, many learning-based AI studies deploy deep neural networks (DNN) with a deep learning (DL) algorithm that implements stochastic gradient descent optimization (SGDO) \cite{amari1993backpropagation}. DNN models use complex network architectures that include many layers and nodes to learn highly nonlinear correlations among data. These DNNs have revolutionized the business and technology world with good performance in areas as varied as image classification, object detection and tracking, and video analytics. The frameworks of TensorFlow \cite{abadi2016tensorflow, abadi2016tensorflow2} or PyTorch \cite{paszke2019pytorch} and graphics processing units (GPUs) are used to enhance the learning performance of DL. However, complications still arise with DL, including long training times, overfitting, and the need for manual hyperparameter tuning \cite{bengio2013representation, lecun2015deep}.

Various attempts \cite{tsaih1993softening, tsaih1998explanation, tsai2019cramming} have been made to derive new learning approaches to resolve overfitting and vanishing gradients. For example, Tsaih \cite{tsaih1993softening, tsaih1998explanation} adopts a hybrid approach to systematically construct and prune hidden nodes of 2LNN with the tanh activation function. \cite{tsai2019cramming} further revise this approach using the cramming, softening, and integrating (CSI) learning algorithm shown in Table \ref{CSI} with parametric rectified linear unit (ReLU) activations instead of tanh activations. In addition, a variety of recent studies apply different methods to resolve the aforementioned issues \cite{lu2017adaptive, ma2019adaptive, niu2018adaptive, wang2020multiscale}. However, most of the studies provide little empirical validation, making the proposed approaches controvertible.

\begin{table}
\renewcommand{\arraystretch}{1.3}
\begin{center}
\caption{CSI (cramming, softening, and integrating) learning algorithm \cite{tsai2019cramming}.}
\label{CSI}
\begin{tabular}{p{8.2cm}}
\hline
\begin{enumerate}
    \item[Step 1:] {
        Use two reference observations $\{(\boldsymbol{\rm x}^1, y^1), (\boldsymbol{\rm x}^2, y^2)\}$ with $y^1 \cdot y^2=-1$ to set up an acceptable SLFN estimate with one hidden node. Set $n=3$.
    }
    \item[Step 2:] {
        If $n > N$, STOP.
    }
    \item[Step 3:] {
        Pick up the first n reference observations $\{(\boldsymbol{\rm x}^c, y^c)\}$ which are sorted by all $N$ reference observations' squared residuals in ascending order. Let $\boldsymbol{\rm I}(n)$ be the set of indices of these picked observations.
    }
    \item[Step 4:] {
        If the condition $L$ regarding $\{f(\boldsymbol{\rm x}^c, \boldsymbol{\rm w}, r), \forall c \in \boldsymbol{\rm I}(n)\}$ is satisfied, go to Step 7; otherwise, there is one and only $\kappa \in \boldsymbol{\rm I}(n)$ that is not at the right place.
    }
    \item[Step 5:] {
        Save $\boldsymbol{\rm w}$ and $r$.
    }
    \item[Step 6:] {
        Apply the weight-tuning mechanism to $\underset{\boldsymbol{\rm w}, r}{\text{min}} E_n(\boldsymbol{\rm w}, r)$ and adjust $\boldsymbol{\rm w}$ and $r$ until one of the following two cases occurs:
        \begin{enumerate}
        \setlength{\itemindent}{-.1in}
            \item[1)] {
                If the condition $L$ regarding $\{f(\boldsymbol{\rm x}^c, \boldsymbol{\rm w}, r), \forall c \in \boldsymbol{\rm I}(n)\}$ is satisfied, go to Step 7.
            }
            \item[2)] {
                If the condition $L$ is not satisfied, restore $\boldsymbol{\rm w}$ and $r$ then apply the cramming mechanism to add one extra hidden node to the existing SLFM to obtain an acceptable SLFN estimate.
            }
        \end{enumerate}
    }
    \item[Step 7:] {
        Apply the softening and integrating mechanism to prune the irrelevant hidden node, $n+1 \to n$; go to Step 2.
    }
\end{enumerate}\\
\hline
\end{tabular}
\end{center}
\end{table}

Extending \cite{tsaih1993softening, tsaih1998explanation, tsai2019cramming} and following the pupil learning procedure shown in Table \ref{plp}, in this study we derive the PLM, which applies to 2LNNs with ReLU activations \cite{hara2015analysis, xu2015empirical} and contains modules for sequential learning, adaptive learning, perfect learning, and less-overfitted learning. Due to the variety of advanced modules developed to process data of high dimensions and large nonlinearity, it is mathematically infeasible to validate the proposed PLM. Therefore, in this study, we conduct a copper price forecasting experiment to empirically validate the PLM module design. We also conduct a copper price forecasting experiment to empirically evaluate the performance of PLM. The empirical results indeed exhibit incremental learning, adaptive learning, perfect learning, and less-overfitted learning as well as that the PLM can effectively (1) adjusts the number of used hidden nodes according to the consequence of learning and new data, and (2) reduces the overfitting tendency while learning all training instances. The empirical results also show the superiority of the proposed PLM model over the linear regression model and the conventional backpropagation-based 2LNN model. 

The rest of the paper is organized as follows: Section 2 describes in detail the proposed PLM. Section 3 presents the experimental design, and Section 4 presents the empirical results. Section 5 concludes and suggests future work.

\section{Proposed Pupil Learning Mechanism}
\noindent Without loss of generalization, we consider the regression problem with real-number inputs and use a 2LNN with one output node defined in Eqs. (\ref{eq1}) and (\ref{eq2}). Table \ref{notation} lists the notation.
\begin{equation}
\label{eq1}
    a_i(\boldsymbol{\rm x}^c, \boldsymbol{\rm w}^H_i) \equiv \text{ReLU} \left(w^H_{i0} + \sum\limits_{j=1}^m{w^H_{ij}x^c_{j}}\right)
\end{equation}

\begin{equation}
\label{eq2}
    f(\boldsymbol{\rm x}^c, {\rm w}) \equiv w^o_0 + \sum\limits_{i=1}^p{w^o_i\text{ReLU} \left(w^H_{i0} + \sum\limits_{j=1}^m{w^H_{ij}x^c_{j}}\right)}
\end{equation}

\begin{table}
\renewcommand{\arraystretch}{1.3}
\begin{center}
\caption{Notation list. Characters in bold represent column vectors, matrices, or sets; superscript $H$ indicates hidden layer quantities; superscript $o$ indicates output layer quantities; $(\cdot)^{\rm T}$ denotes the transpose of $(\cdot)$.}
\label{notation}
\begin{tabular}{p{8.2cm}}
\hline
    $\text{ReLU}(x) \equiv \text{max}(0, x);$\\   
    $m$: the number of input nodes;\\
    $\boldsymbol{\rm x}^c \equiv (x^c_1, x^c_2, \cdots, x^c_m)^T$: the $c^{\rm th}$ input;\\
    ${y}^c$: the desired output of the $c^{\rm th}$ input;\\
    $N$: the total amount of training instances;\\
    $p$: the number of adopted hidden nodes;\\
    $w^H_{i0}$: the bias of $i^{\rm th}$ hidden nodes;\\
    $w^H_{ij}$: the weight between the $j^{\rm th}$ input variable and the $i^{th}$ hidden nodes;\\
    $\boldsymbol{\rm w}^H_i \equiv (w^H_{i0}, w^H_{i1}, w^H_{i2}, \cdots w^H_{im})^{\rm T}$;\\
    $w^o_0$: the bias value of the output node;\\
    $w^o_i$: the weight between the $i^{\rm th}$ hidden node and the output node;\\
    $\boldsymbol{\rm w}^o \equiv (w^o_{0}, w^o_{1}, w^o_{2}, \cdots , w^o_{p})^{\rm T}$\\
    $\boldsymbol{\rm w}^H \equiv (\boldsymbol{\rm w}^H_{0}, \boldsymbol{\rm w}^H_{1}, \boldsymbol{\rm w}^H_{2}, \cdots , \boldsymbol{\rm w}^H_{p})^{\rm T}$;\\
    $\boldsymbol{\rm w} \equiv \{ \boldsymbol{\rm w}^o, \boldsymbol{\rm w}^H\}$;\\
    $e^c \equiv f(\boldsymbol{\rm x}^c, \boldsymbol{\rm w}) - y^c$.\\
\hline
\end{tabular}
\end{center}
\end{table}

\begin{figure}[!t]
\centering
\includegraphics[width=\linewidth]{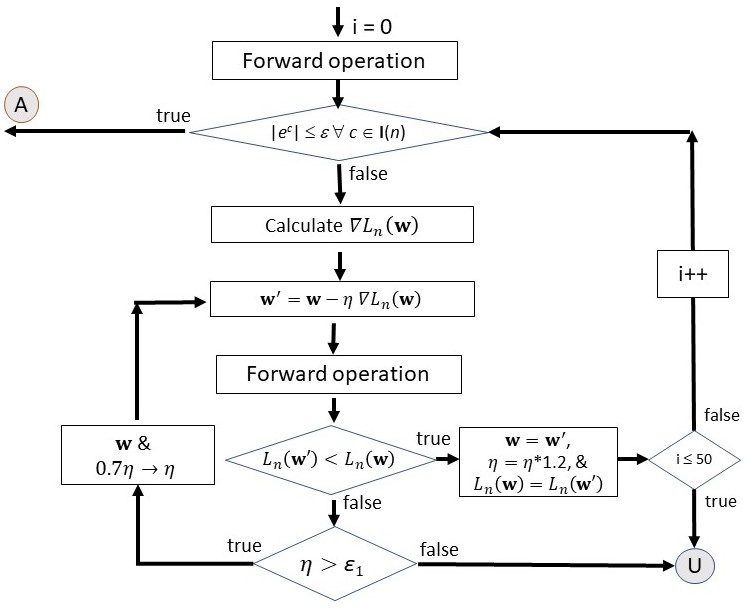}
\caption{AGDO implementing an optimizer with an adaptive learning rate $\eta$.}
\label{Fig:AGDO}
\end{figure}

At the n-th stage, let $\boldsymbol{\rm I}(n)$ be the subset of indices of training instances. The learning goal is to find $w$ where
\begin{equation*}
    \lvert e^c\rvert \leq \varepsilon, \forall c \in \boldsymbol{\rm I}(n).
\end{equation*}
In this study, the c-th training instance is acceptable if $\lvert e^c\rvert \leq \varepsilon$; otherwise, it is unacceptable. Fig. \ref{Fig:AGDO} shows the adaptive GDO (AGDO) implementing an optimizer (e.g., the Adam optimizer \cite{kingma2014adam}) with automatic adjustment of learning rate $\eta$ to calculate $\nabla L_n(\boldsymbol{\rm w})$, where
\begin{equation}
\label{AGDO:loss}
    L_n(\boldsymbol{\rm w}) \equiv \sum_{c \in \boldsymbol{\rm I}(n)}{\left(e^c \right)}^2/n.
\end{equation}
However, AGDO does not eliminate vanishing gradients and thus may yield an unacceptable 2LNN due to convergence at a local minimum or a saddle point of $L_n(\boldsymbol{\rm w})$. Rate $\eta$ is smaller than $\varepsilon_1$ (a tiny value) and the upper iteration bound $(i \leq 50)$ is designed to identify unacceptable 2LNNs. In Fig. \ref{Fig:AGDO}, exit point $A$ denotes an acceptable 2LNN, and $U$ denotes an unacceptable 2LNN.

\begin{figure}[!t]
\begin{center}
\includegraphics[width=\linewidth]{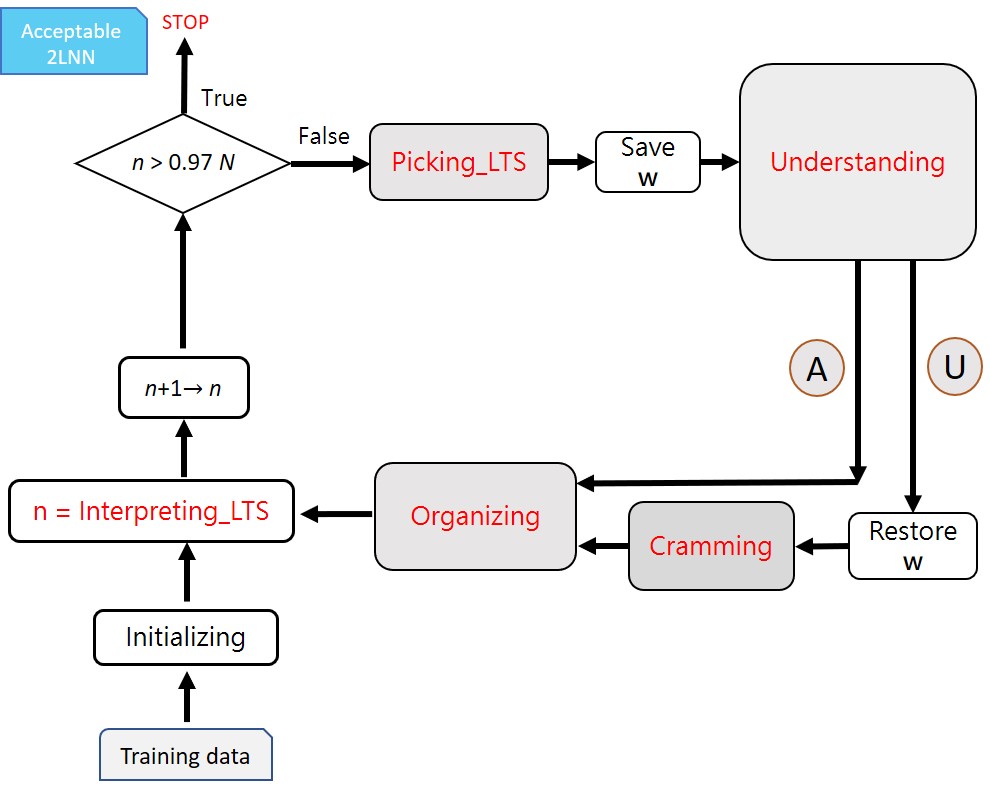}
\end{center}
\caption{Flow chart of proposed PLM.}
\label{Fig:PLM}
\end{figure}

A flowchart for the proposed PLM is shown in Fig. \ref{Fig:PLM}. Details are provided as follows. Let
\begin{equation*}
    D_n = \underset{c \in \boldsymbol{\rm I}(n)}{\text{max}} \lvert f(\boldsymbol{\rm x}^c, \boldsymbol{\rm w}) - y^c \rvert,
\end{equation*}
where a 2LNN produces output $f(\boldsymbol{\rm x}^c, \boldsymbol{\rm w})$ when input is $\boldsymbol{\rm x}^c$ and the desired output is $y^c$.

The initializing module sets up a 2LNN with one hidden node and then uses the GDO to tune the weights and thresholds of the 2LNN with the training instances $\{(\boldsymbol{\rm x}^c, y^c): c \in \boldsymbol{\rm I}(N) \}$. Least trimmed squares (LTS) \cite{tsaih2009resistant, tsai2019cramming} is defined as minimizing $\sum_{c=1}^q(e^{[c]})^2$, where only the $q$ smallest ordered squared residual values are included in the summation, and $(e^{[c]})^2$ denotes the sorted squared residuals in ascending order $(e^{[1]})^2 \leq (e^{[2]})^2 \leq \cdots \leq (e^{[N]})^2$. The interpreting\_LTS module implements the LTS principle, which first sorts all training instances by their squared residuals in ascending order as$\left(e^{[1]}\right)^2 \leq \left(e^{[2]}\right)^2 \leq \cdots \leq \left(e^{[N]}\right)^2$. Second, the interpreting\_LTS module examines the acceptability of every instance (i.e., the $c^{\rm th}$ instance is acceptable if $\lvert e^c \leq \varepsilon$). Third, the interpreting\_LTS module yields the $n$ value, the number of acceptable instances. Thus, $D_n \leq \varepsilon$ after the interpreting\_LTS module and before the sequential module $(n+1 \to n)$.

The stopping criterion $(n < 0.97N)$ associated with the sequential module indicates that the PLM sequentially learns the training instances until more than $\lfloor 0.97N \rfloor$ instances are acceptable, where $\lfloor x \rfloor$ is the largest integer less than or equal to $x$. In other words, the PLM learns merely the majority of training data (with at most $\lfloor 0.97N \rfloor$ data), but the learning terminates when the resultant 2LNN renders more than $\lfloor 0.97N \rfloor$ (training) data acceptable. 

For $n < 0.97N$, the picking\_LTS module selects the first n instances with the smallest squared residuals as the training instances, for which $\boldsymbol{\rm I}(n)$ is the set of instance indices. Note that the $\boldsymbol{\rm I}(n)$ subset contains merely one unacceptable instance: the $[n]$-th instance. This prevents the PLM not only from learning instances with much larger squared residuals that are difficult for the learning mechanism to understand, but also from causing more cramming occurrences, which may lead to too many hidden nodes and thus overfitting. Then, $\boldsymbol{\rm w}$ is saved such that when $\boldsymbol{\rm w}$ is restored we can resume the 2LNN estimate to the scenario in which there is only one unacceptable instance in the $\boldsymbol{\rm I}(n)$ subset. The interpreting\_LTS module and picking\_LTS modules together yield a learning sequence in which easy instances are learned first.

The understanding module implements the AGDO of Fig. \ref{Fig:AGDO} with the loss function defined at equation \ref{AGDO:loss} to learn all training instances of $\boldsymbol{\rm I}(n)$. If the understanding module results in an acceptable 2LNN estimate, then $D_n \leq \varepsilon$. Otherwise, the PLM restores $\boldsymbol{\rm w}$ and triggers the cramming module of Table \ref{cramming module} to process instance $[n]$ by adding three hidden nodes so the 2LNN now has $p$ hidden nodes. That is, with the restore module,
\begin{equation*}
    \left(y^{[n]} - w^o_0 - \sum^{p-3}_{i=1}{w^o_i a^{[n]}_i}\right)^2 > \varepsilon^2,
\end{equation*}
\begin{equation*}
    \left(y^{c} - w^o_0 - \sum^{p-3}_{i=1}{w^o_i a^{c}_i}\right)^2 \leq \varepsilon^2, \forall c \in \boldsymbol{\rm I}(n) \setminus \{[n]\}.
\end{equation*}

\begin{table}
\renewcommand{\arraystretch}{2}
\begin{center}
\caption{Cramming Module}
\label{cramming module}
\begin{tabular}{p{8.2cm}}
\hline
    \begin{enumerate}
    \item[Step 1:] {
        Select a tiny positive number $\zeta$ and randomly generate an $m$-vector $\boldsymbol{\gamma}$ of length one such that 
        \begin{equation*}
            \boldsymbol{\gamma}^T\left(\boldsymbol{\rm x}^c - \boldsymbol{\rm x}^{[n]}\right) \neq 0,
        \end{equation*}
        \begin{multline*}
            \left(\zeta + \boldsymbol{\gamma}^T\left(\boldsymbol{\rm x}^c - \boldsymbol{\rm x}^{[n]}\right)\right)\left(\zeta - \boldsymbol{\gamma}^T \left(\boldsymbol{\rm x}^c - \boldsymbol{\rm x}^{[n]}\right)\right) < 0,\\
            \forall c \in \boldsymbol{\rm I}(n) \setminus \{[n]\}.
        \end{multline*}
    }
    \item[Step 2:] {
        Let $p+3 \to p$, add three new hidden nodes $p-2$, $p-1$, and $p$ to the current 2LNN, and assign their associated weights and thresholds as follows to ensure that $\lvert e^c \rvert \leq \varepsilon, \forall c \in \boldsymbol{\rm I}(n)$ is true:
        \begin{equation*}
             \boldsymbol{\rm w}^H_{p-2} = \boldsymbol{\gamma}, 
            w^H_{p-2, 0} = \zeta - \boldsymbol{\gamma}^T\boldsymbol{\rm x}^{[n]},
        \end{equation*}
        \begin{equation*}
            w^o_{p-2} = \frac{y^{[n]} - w^o_0 - \sum_{i=1}^{p-3}{w^o_i a^{[n]}_i}}{\zeta},
        \end{equation*}
            
        \begin{equation*}
            \boldsymbol{\rm w}^H_{p-1} = \boldsymbol{\gamma}, 
            w^H_{p-1, 0} = - \boldsymbol{\gamma}^T\boldsymbol{\rm x}^{[n]},
        \end{equation*}
        \begin{equation*}
            w^o_{p-1} = \frac{-2 \left( y^{[n]} - w^o_0 - \sum_{i=1}^{p-3}{w^o_i a^{[n]}_i} \right) }{\zeta},
        \end{equation*}
        
        \begin{equation*}
            \boldsymbol{\rm w}^H_{p} = \boldsymbol{\gamma}, 
            w^H_{p, 0} = - \zeta - \boldsymbol{\gamma}^T\boldsymbol{\rm x}^{[n]},
        \end{equation*}
        \begin{equation*}
            w^o_{p} = \frac{y^{[n]} - w^o_0 - \sum_{i=1}^{p-3}{w^o_i a^{[n]}_i}}{\zeta}.
        \end{equation*}
    }
\end{enumerate}\\
\hline
\end{tabular}
\end{center}
\end{table}

To compute $D_n$, we first consider instance $[n]$ as input and show that the output of 2LNN, $f(\boldsymbol{\rm x}^{[n]}, \boldsymbol{\rm w})$, equals exactly the desired output $y^{[n]}$. By Step 2 of the cramming module,
\begin{align*}
    a^{[n]}_{p-2} &= \text{ReLU} \left( w^H_{p-2,0} + \sum^m_{j=1}{w^H_{p-2,j}x^{[n]}_j} \right)\\
    &=  \text{ReLU} \left( \zeta + \boldsymbol{\gamma}^T\left(\boldsymbol{\rm x}^{[n]} - \boldsymbol{\rm x}^{[n]}\right) \right)\\
    &= \zeta.
\end{align*}

Similarly, 

\begin{equation*}
a^{[n]}_{p-1} = \text{ReLU} \left( \boldsymbol{\gamma}^T \left( \boldsymbol{\rm x}^{[n]} - \boldsymbol{\rm x}^{[n]}\right) \right)
= 0,
\end{equation*}

\begin{equation*}
a^{[n]}_{p} = \text{ReLU} \left( -\zeta + \boldsymbol{\gamma}^T \left( \boldsymbol{\rm x}^{[n]} - \boldsymbol{\rm x}^{[n]}\right) \right) = 0.    
\end{equation*}

Since $f(\boldsymbol{\rm x}^{[n]}, \boldsymbol{\rm w})$ equals the contribution of the first $p-3$ hidden nodes to the output node plus contribution of the $p-2^{th}$, $p-1^{th}$ and $p^{th}$ hidden nodes to the output node, by the assignment of $w^o_{p-2}$, $w^o_{p-1}$, and $w^o_{p}$ in Step 2 of the cramming module,
\begin{align*}
    f(\boldsymbol{\rm x}^{[n]}, \boldsymbol{\rm w}) &= w^o_0 + \sum^{p-3}_{i=1}w^o_ia^{[n]}_i\\
    &+ \zeta \left( \frac{y^[n] - w^o_0 - \sum^{p-3}_{i=1}w^o_ia^{[n]}_i}{\zeta} \right)\\
    &= y^{[n]}
\end{align*}

Thus, when instance $[n]$ is input and the cramming module is executed, no additional effect on $D_n$ is introduced by instance $[n]$.

With regard to the effect on $D_n$ of instances $c \in \boldsymbol{\rm I}(n) \setminus\{[n]\}$, we argue that these instances will not increase $D_n$. The change in the output node is due to the inputs from the $p-2^{th}$, $p-1^{th}$ and $p^{th}$ hidden nodes. If instance $c$ is input, 
\begin{align*}
     a^c_{p-2} &= \text{ReLU}\left( w^H_{p-2, 0} + \sum^m_{j=1}{w^H_{p-2, j} x^c_j}\right)\\
     &= \text{ReLU}\left( \zeta + \boldsymbol{\gamma}^T\left(\boldsymbol{\rm x}^c - \boldsymbol{\rm x}^{[n]}\right) \right). 
\end{align*}

Similarly,
\begin{equation*}
   a^c_{p-1}=\text{ReLU}\left( \boldsymbol{\gamma}^T\left(\boldsymbol{\rm x}^c - \boldsymbol{\rm x}^{[n]}\right) \right),
\end{equation*}
\begin{equation*}
    a^c_{p-1} = \text{ReLU}\left( -\zeta + \boldsymbol{\gamma}^T\left(\boldsymbol{\rm x}^c - \boldsymbol{\rm x}^{[n]}\right) \right).
\end{equation*}

By the assignment of $w_{p-2}^o$, $w_{p-1}^o$, and $w_p^o$ in Step 2 of the cramming module,
\begin{align*}
    \sum_{i=p-2}^p{w^o_{i}a^c_{i}} &= w^o_p \Bigl[\text{ReLU}\left( \zeta + \boldsymbol{\gamma}^T\left(\boldsymbol{\rm x}^c - \boldsymbol{\rm x}^{[n]}\right) \right)\\
    &- 2\text{ReLU}\left( \boldsymbol{\gamma}^T\left(\boldsymbol{\rm x}^c - \boldsymbol{\rm x}^{[n]}\right) \right)\\
    &+ \text{ReLU}\left( -\zeta + \boldsymbol{\gamma}^T\left(\boldsymbol{\rm x}^c - \boldsymbol{\rm x}^{[n]}\right) \right)\Bigr] \\
    &= g(c).
\end{align*}

In Step 1 of the Table \ref{cramming module}, we have set $\zeta$ and $\boldsymbol{\gamma}$; therefore, $g(c) = 0, \forall c \in \boldsymbol{\rm I}(n) \setminus \{[n]\}$. Thus, the cramming module makes the change in the output node in all instances $c \in \boldsymbol{\rm I}(n) \setminus \{[n]\}$ equal to zero. In other words, $D_n \leq \varepsilon$, and by adding three extra hidden nodes with the ReLU activation function, the cramming module renders $\lvert e^c \rvert \leq \varepsilon, \forall c \in \boldsymbol{\rm I}(n)$ true immediately.

\begin{figure}[!t]
\begin{center}
\includegraphics[width=\linewidth]{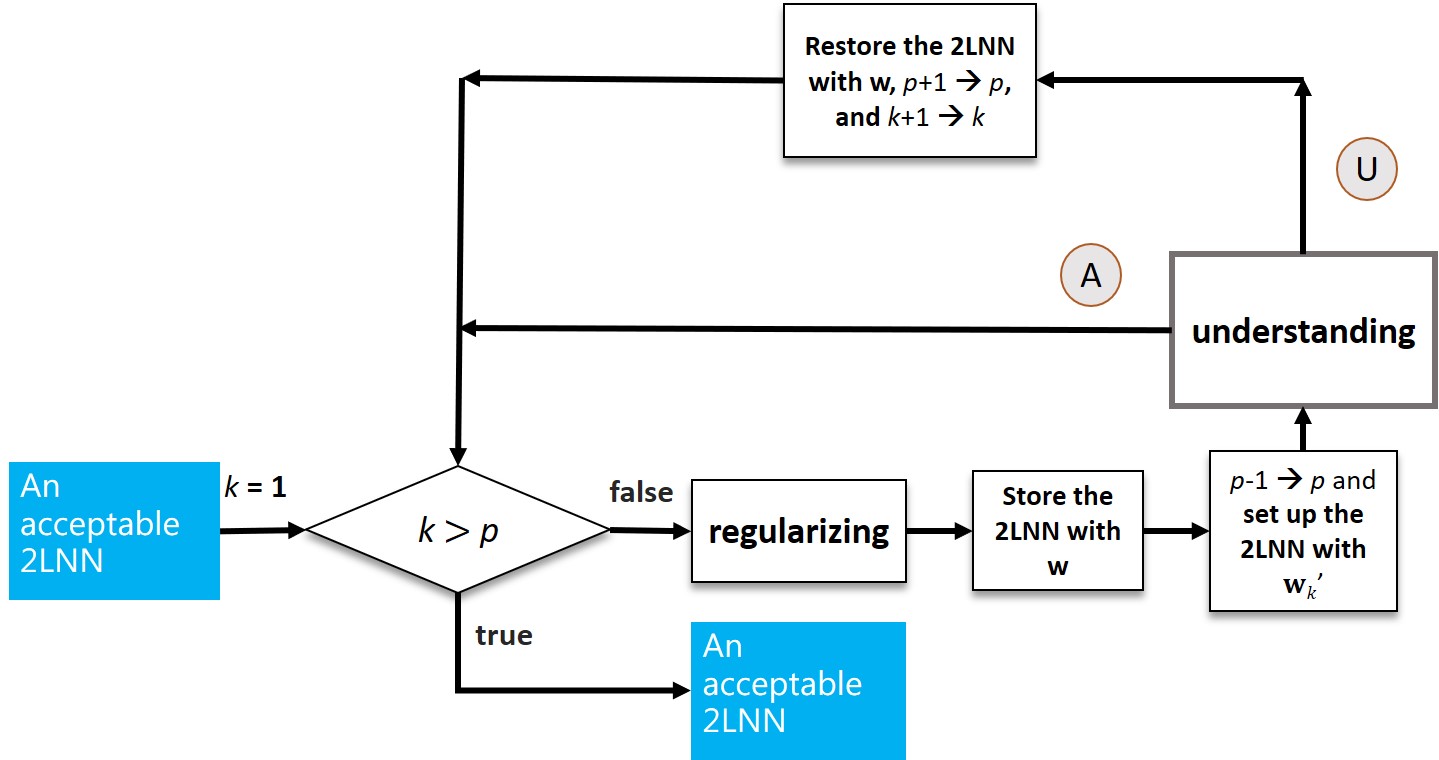}
\end{center}
\caption{Organizing module where $\boldsymbol{\rm w}^{\prime}_k \equiv \boldsymbol{\rm w} \setminus \{ w^0_k, \boldsymbol{\rm w}^H_k \}$.}
\label{Fig:organizing}
\end{figure}

\begin{figure}[!t]
\begin{center}
\includegraphics[width=\linewidth]{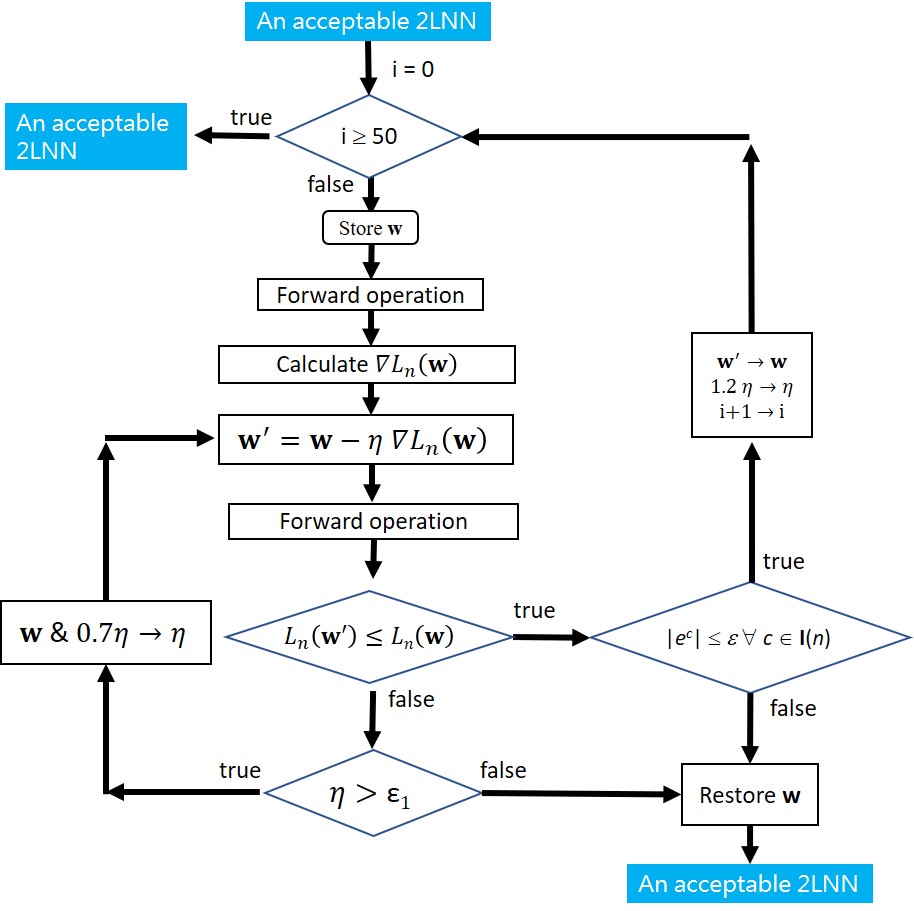}
\end{center}
\caption{Regularizing module.}
\label{Fig:regularizing}
\end{figure}

The organizing module of Fig. \ref{Fig:organizing} reduces overfitting by identifying and pruning irrelevant hidden nodes \cite{tsaih1998explanation}. As shown in Fig. \ref{Fig:organizing}, the organizing module first implements the regularizing module of Fig. \ref{Fig:regularizing}, 
which regularizes the weights and thresholds by means of the loss function
\begin{equation}
\label{reg loss}
    Lr_n(\boldsymbol{\rm w}) \equiv \sum_{c \in \boldsymbol{\rm I}(n)}{(e^c)^2} / n + \lambda {\lVert \boldsymbol{\rm w} \rVert}^2,
\end{equation}
where $\lambda$ is the regularization coefficient. In contrast with the AGDO of Fig. \ref{Fig:AGDO}, the regularizing module deploys AGDO to tune $\boldsymbol{\rm w}$ while keeping $\lvert e^c \rvert \leq \varepsilon, \forall c \in \boldsymbol{\rm I}(n)$ true. Namely, with the loss function defined at equation \ref{reg loss}, the regularizing module attempts to prevent large-magnitude weights and thresholds while keeping $\lvert e^c \rvert \leq \varepsilon, \forall c \in \boldsymbol{\rm I}(n)$ true. Thus $D_n \leq \varepsilon$ after implementing the regularizing module.

After implementing the regularizing module, the organizing module stores $\boldsymbol{\rm w}$ and creates a new 2LNN with $p-1 \to p$, the $k^{th}$ hidden node is removed, and the weights and thresholds are assigned as $\boldsymbol{\rm w}_k^{\prime} \equiv \boldsymbol{\rm w} \setminus \{ w^o_k, \boldsymbol{\rm w}^H_k \}$. Then, the understanding module with the AGDO of Fig. \ref{Fig:AGDO} is applied to this new 2LNN to $\underset{\boldsymbol{\rm w}_k^{\prime}}{\text{min}}\left(L_n \left( \boldsymbol{\rm w}_k^{\prime}\right)\right)$ and $\boldsymbol{\rm w}_k^{\prime}$ is adjusted with the loss function at equation \ref{AGDO:loss}. If the understanding module yields an acceptable 2LNN, the afore-mentioned process is applied again to the next hidden node. If the understanding module yields an unacceptable 2LNN, then the stored 2LNN with $\boldsymbol{\rm w}$ is restored, $p+1 \to p, k+1 \to k$, and the afore-mentioned process is again applied to the next hidden node.

The organizing module ensures that $\lvert e^c\rvert \leq \varepsilon, \forall c \in \boldsymbol{\rm I}(n)$ is true. Thus $D_n \leq \varepsilon$ and we have the following Lemma. In addition, the organizing module helps to identify and prune irrelevant hidden nodes to reduce overfitting in the resulting 2LNN.

\begin{lemma}
For all $n$, the PLM yields a 2LNN with $D_n \leq \varepsilon$.
\end{lemma}
Note that when learning a new (unacquainted) instance, the PLM results in an acceptable 2LNN via one of the following two routes:
\begin{enumerate}
    \item {
        Understanding route by implementing the AGDO of Fig. \ref{Fig:AGDO} to yield an acceptable 2LNN.
    }
    \item {
        Cramming route by implementing the cramming module of Table \ref{cramming module} to yield an acceptable 2LNN.
    }
\end{enumerate}

\section{Experiment Design}
\noindent To validate the proposed PLM, we conducted an experiment on copper price forecasting to determine the following: 1) whether the understanding and cramming routes are effective; 2) whether the sequential module and the cramming module effectively address the vanishing gradient problem; and 3) whether LTS and the organizing arrangements reduce overfitting. This study further examines whether the total PLM training time is reasonable and whether the PLM yields better forecast accuracy than other tools mentioned in the literature.
Based upon the literature review regarding copper price forecasting, this study identifies the 18 input variables and the four-week-ahead-forecast output variables shown in Table \ref{IO}. Therefore, $m = 18$.

\begin{table}
\renewcommand{\arraystretch}{1.3}
\begin{center}
\caption{Input and output variables.}
\label{IO}
\begin{tabular}{c p{5cm} c}
\hline
   Variable         &   \centering Description  &   Ref.  \\
   \hline
   $x^t_1$          &   Weekly crude oil price of New York Mercantile Exchange at time epoch $t$ &   \cite{alameer2019forecasting, liu2017forecasting}     \\
   $x^t_2$          &   Weekly copper spot price of YR nonferrous metals at epoch $t$   &        \\
   $x^t_3$          &   Weekly copper spot price of YR nonferrous metals at epoch $t-1$ &        \\
   $x^t_4$          &   Weekly copper spot price of YR nonferrous metals at epoch $t-2$ &        \\
   $x^t_5$          &   Weekly copper spot price of YR nonferrous metals at epoch $t-3$ &        \\
   $x^t_6$          &   Weekly copper spot price of London Metal Exchange at epoch $t$  &   \cite{alameer2019forecasting}      \\
   $x^t_7$          &   Weekly gold spot price of FX Broker at epoch $t$                &   \cite{alameer2019forecasting}      \\
   $x^t_8$          &   Weekly silver spot price of FX Broker at epoch $t$              &   \cite{alameer2019forecasting}      \\
   $x^t_9$          &   Weekly nickel spot price of London Metal Exchange at epoch $t$  &   \cite{alameer2019forecasting}      \\
   $x^t_{10}$       &  Weekly aluminum spot price of London Metal Exchange at epoch $t$ &   \cite{alameer2019forecasting}      \\
   $x^t_{11}$       &  Weekly zinc spot price of London Metal Exchange at epoch $t$     &   \cite{alameer2019forecasting}      \\
   $x^t_{12}$       &  Weekly iron spot price of London Metal Exchange at epoch $t$     &   \cite{alameer2019forecasting}      \\
   $x^t_{13}$       &  US inflation rates at epoch $t$                                  &   \cite{alameer2019forecasting, orlowski2017volatility}      \\
   $x^t_{14}$       &  China inflation rates at epoch $t$                               &   \cite{alameer2019forecasting, orlowski2017volatility}      \\
   $x^t_{15}$       &  Weekly USD/CLP dollar exchange rate at epoch $t$                 &   \cite{alameer2019forecasting, wang2019spillover}\\
   $x^t_{16}$       &  Weekly USD/PEN dollar exchange rate at epoch $t$                 &   \cite{alameer2019forecasting, wang2019spillover}\\
   $x^t_{17}$       &  Weekly USD/RMB dollar exchange rate at epoch $t$                 &   \cite{alameer2019forecasting, wang2019spillover}\\
   $x^t_{18}$       &  Weekly USD/EURO dollar exchange rate at epoch $t$                &   \cite{alameer2019forecasting, wang2019spillover}\\
   $y^t$            &  Weekly copper spot price of YR nonferrous metals at epoch $t+4$  &   \cite{astudillo2020copper}     \\
\hline
\end{tabular}
\end{center}
\end{table}

The dataset includes 471 weekly copper prices of Yangtze River (YR) nonferrous metals from 2011/10/31 to 2020/12/21. The average (AVG) and standard deviation (SD) of $y$ are 48,358.75 RMB/ton and 6,183.97 RMB/ton, respectively. In this study, $y$ is divided by 100,000 and its values fall between 0 and 1. Twenty datasets were generated by randomly picking 282 instances, which comprised approximately 60\% of the dataset of 471 instances, as the training data and the remaining 189 instances were testing data. Therefore, $N = 282$ and the learned majority of training data was at most 272 data.

To explore the study objectives, we used four different versions of the PLM shown in Table \ref{PLM version}. $LTS^N_n$ indicates that both the interpretating module and the picking module follow the LTS principle to yield the $n$ value, which is set to the count of acceptable instances, and pick the first $n$ instances, respectively, whereas $PO^N_n$ denotes that both the interpretating module and the picking module follow the pre-ordered principle. Organizing($x$) indicates that the organizing module regularizes weights for at most $x$ epochs. These versions were implemented with the PyTorch framework and GPU hardware shown in Table \ref{computer env}.

\begin{table}
\renewcommand{\arraystretch}{1.3}
\caption{PLM Versions and Settings.}
\label{PLM version}
\centering
\begin{tabular}{c p{2.8cm} c}
\hline
   Version &   \centering Interpreting module and picking module & Organizing module\\
\hline
   PLM-PO-100   &   \centering$PO^N_n$   &   organizing(100)   \\
   PLM-LTS-0    &   \centering$LTS^N_n$  &   organizing(0)     \\
   PLM-LTS-100  &   \centering$LTS^N_n$  &   organizing(100)   \\
   PLM-LTS-500  &   \centering$LTS^N_n$  &   organizing(500)   \\
\hline
\end{tabular}
\end{table}

\begin{table}
\renewcommand{\arraystretch}{1.3}
\caption{Computational Environment.}
\label{computer env}
\centering
\begin{tabular}{cc}
\hline
   OS                     &  Ubuntu 20.0.4.4 LTS            \\
   Programming Language   &  Python 3.7.7                   \\
   Pytorch version        &  1.11.0                         \\
   IDE                    &  PyCharm, Jupyter Notebook      \\
   GPU                    &  GeForce RTX 2080               \\
   RAM                    &  DDR-4 8G*4                     \\
\hline
\end{tabular}
\end{table}

The $\varepsilon$ value for the interpretating, understanding, and organizing modules is set to 0.04836 (i.e., 10\% of the average of the normalized $y$). Thus, when $-0.04836 \leq f\left( \boldsymbol{\rm x}^c, \boldsymbol{\rm w} \right) - y^c \leq 0.04836$, the $c^{th}$ instance is classified as acceptable; otherwise, it is unacceptable. The initial learning rates $\eta$ in the understanding and organizing modules are set to $1e-2$ and $1e-3$, respectively. Both understanding and organizing modules have an $\varepsilon_1$ value of $1e-7$.

\section{Experimental Results}
\begin{figure}[!t]
\begin{center}
\includegraphics[width=\linewidth]{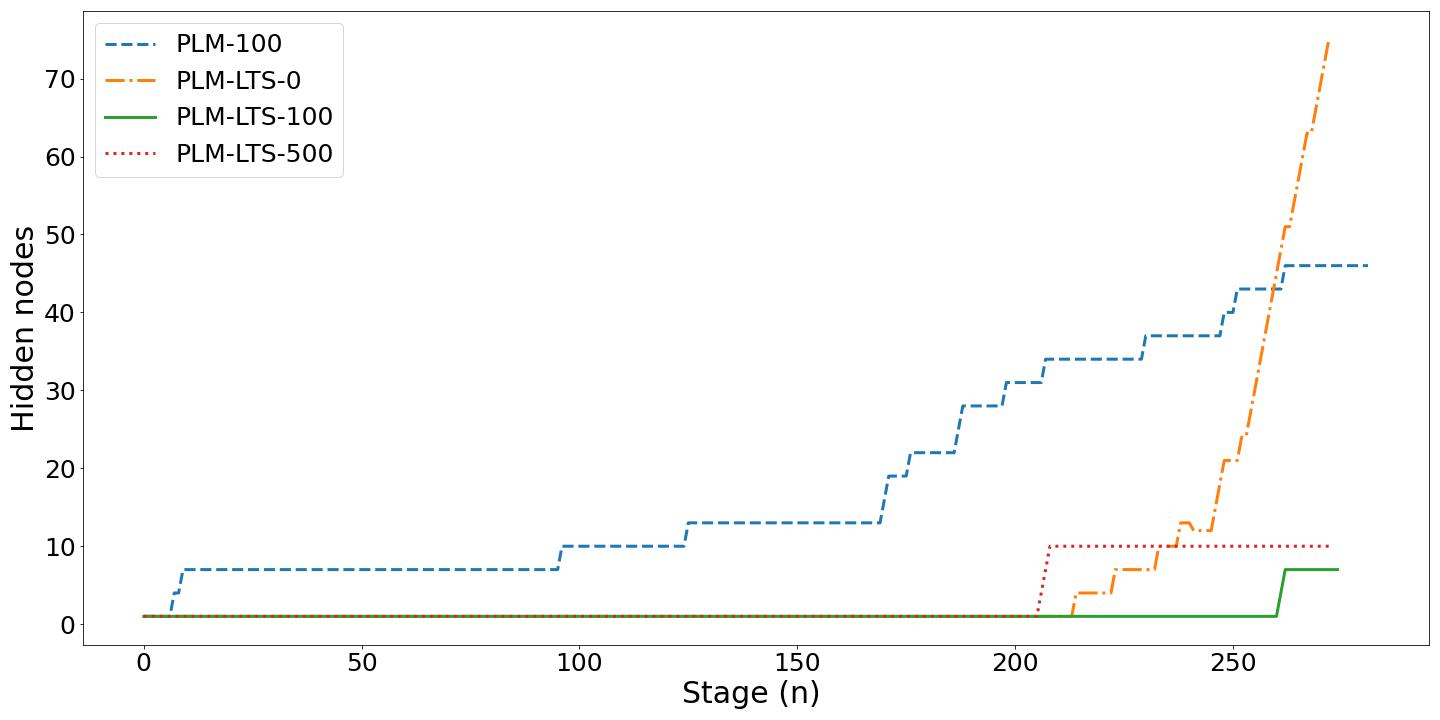}
\end{center}
\caption{Number of hidden nodes during learning process of $10^{th}$ training dataset.}
\label{Fig:hidden node 10}
\end{figure}

\noindent Fig. \ref{Fig:hidden node 10} shows the evolution of the number of hidden nodes during the learning process of the $10^{th}$ training dataset for the four versions. As expected, cramming was triggered several times in the early stage of the learning process of PLM-PO-100. In contrast, in the early stage of the learning process of PLM-LTS-100 (and other PLM-LTS-xxx versions), the cramming module was not often triggered. Furthermore, compared with the learning process of PLM-LTS-0 shown in Fig. \ref{Fig:hidden node 10} with those of PLM-LTS-100 and PLM-LTS-500, the non-zero regularizing epochs reduce the frequency of cramming.

Tables \ref{understanding route} and \ref{cramming route} show the occurrence frequencies of the understanding route and cramming route over twenty training datasets for the four versions. All non-zero minimal occurrence frequencies of the cramming route for PLM-PO-100, PLM-LTS-0, and PLM-LTS-100 shown in Table \ref{cramming route} indicate an unavoidable vanishing gradient problem for the understanding module. The cramming module is indeed required to address the vanishing gradient. It is worth noting that a zero-value minimal occurrence frequency of the cramming route for PLM-LTS-500 is shown in Table \ref{cramming route}.

\begin{table}
\renewcommand{\arraystretch}{1.3}
\caption{Occurrence Frequencies with Understanding Route.}
\label{understanding route}
\centering
\begin{tabular}{ccccc}
\hline
          &  PLM-PO-100  &  PLM-LTS-0  &  PLM-LTS-100  &  PLM-LTS-500 \\
\hline
    Min   &  30.41\%     &  10.53\%    &  27.78\%      &  14.29\%     \\
    Max   &  77.14\%     &  77.42\%    &  78.57\%      &  100.00\%    \\
    Avg   &  58.26\%     &  44.31\%    &  56.91\%      &  57.32\%     \\
    SD    &  12.38\%     &  19.67\%    &  13.73\%      &  22.01\%     \\
\hline
\end{tabular}
\end{table}

\begin{table}
\renewcommand{\arraystretch}{1.3}
\caption{Occurrence Frequencies with Cramming Route.}
\label{cramming route}
\centering
\begin{tabular}{ccccc}
\hline
          &  PLM-PO-100  &  PLM-LTS-0  &  PLM-LTS-100  &  PLM-LTS-500 \\
\hline
    Min   &  22.86\%     &  22.58\%    &  21.43\%      &   0.00\%     \\
    Max   &  69.59\%     &  89.47\%    &  72.22\%      &  85.71\%     \\
    Avg   &  41.74\%     &  55.70\%    &  43.09\%      &  42.68\%     \\
    SD    &  12.38\%     &  19.67\%    &  13.73\%      &  22.01\%     \\
\hline
\end{tabular}
\end{table}

Table \ref{end hidden nodes} shows the total number of hidden nodes at the end of learning over twenty training datasets for the four versions. Comparing the PLM-PO-100 and PLM-LTS-100 results, we see that LTS indeed reduces the AVG and SD of the number of hidden nodes and thus reduces overfitting. Comparing the PLM-LTS results, we also see that the more regularizing epochs, the better the minimum, maximum, AVG, and SD of the number of hidden nodes. In other words, regularizing epochs reduce overfitting.

\begin{table}
\renewcommand{\arraystretch}{1.3}
\caption{Hidden Nodes at the end of Learning.}\label{end hidden nodes}
\centering
\begin{tabular}{ccccc}
\hline
          &  PLM-PO-100  &  PLM-LTS-0  &  PLM-LTS-100  &  PLM-LTS-500 \\
\hline
    Min   &  16          &  13         &  4            &  1           \\
    Max   &  301         &  96         &  66           &  55          \\
    Avg   &  52.15       &  44.00      &  29.65        &  22.05       \\
    SD    &  60.93       &  26.92      &  19.24        &  15.83       \\
\hline
\end{tabular}
\end{table}

Table \ref{pruned nodes} shows the number of hidden nodes pruned during the learning process over twenty training datasets for the four versions. It seems that for these four versions, the organizing module indeed works frequently, but does not guarantee successful pruning over a single learning process since the minimal number of hidden nodes pruned within the learning process over twenty training datasets are all zero. In terms of the average number of hidden nodes pruned over the learning process, PLM-LTS-500 has the fewest and PLM-PO-100 has the most.

\begin{table}
\renewcommand{\arraystretch}{1.3}
\caption{Hidden Nodes Pruned over Learning Process.}
\label{pruned nodes}
\centering
\begin{tabular}{ccccc}
\hline
          &  PLM-PO-100  &  PLM-LTS-0  &  PLM-LTS-100  &  PLM-LTS-500 \\
\hline
    Min   &  0           &  0          &  0            &  0           \\
    Max   &  9           &  4          &  4            &  4           \\
    Avg   &  2.10        &  0.95       &  0.90         &  0.79        \\
    SD    &  2.27        &  1.32       &  1.33         &  1.38        \\
\hline
\end{tabular}
\end{table}

Table \ref{training time} shows the training times over twenty training datasets for the four versions. PLM-LTS-500 has the smallest average and standard deviation, and PLM-PO-100 has the largest. Taken together with the results of Tables \ref{end hidden nodes} and \ref{training time}, we observe that the adoption of LTS and the “longer” regularizing module reduce the average and standard deviation of the hidden nodes and thus the training time.

\begin{table}
\renewcommand{\arraystretch}{1.3}
\caption{Training Time (in second).}\label{training time}
\centering
\begin{tabular}{ccccc}
\hline
          &  PLM-PO-100  &  PLM-LTS-0  &  PLM-LTS-100  &  PLM-LTS-500 \\
\hline
    Min   &  51.58       &  50.81      &  13.75        &  12.50       \\
    Max   &  8619.52     &  1753.65    &  1126.95      &  832.77      \\
    Avg   &  766.15      &  589.21     &  430.30       &  222.55      \\
    SD    &  1862.51     &  528.63     &  377.19       &  247.92      \\
\hline
\end{tabular}
\end{table}

Table \ref{MAE PLM} shows the mean absolute error (MAE) over twenty datasets for the four versions. Here, the MAE of the training data (majority) is calculated for the first 272 training data sorted by the LTS principle, the MAE of the training data (non-majority) is calculated for the remaining 10 training data, and the MAE of the testing data is calculated for the 189 testing data. The last row of Table \ref{MAE PLM} displays the ratios of the average MAE of the testing data to the average MAE of the training data (majority), suggesting that the smaller the ratio, the less overfitting there is. Table \ref{MAE PLM} shows that PLM-PO-100 has the highest ratio and PLM-LTS-500 has the smallest ratio. LTS and the “longer” regularizing module appear to reduce overfitting.

\begin{table}
\renewcommand{\arraystretch}{1.3}
\caption{MAE Results.}\label{MAE PLM}
\centering
\begin{tabular}{ccccc}
\hline
          &  PLM-PO-100  &  PLM-LTS-0  &  PLM-LTS-100  &  PLM-LTS-500 \\
\hline
    \multicolumn{5}{c}{Training Data (majority)}                  \\
\hline
    Min   &  0.0097      &  0.0081     &  0.0097       &  0.0092  \\
    Max   &  0.0132      &  0.0156     &  0.0166       &  0.0148  \\
    Avg   &  0.0116      &  0.0119     &  0.0129       &  0.0132  \\
    SD    &  0.0011      &  0.0023     &  0.0016       &  0.0014  \\
\hline
    
    \multicolumn{5}{c}{Training Data (non-majority)}              \\
\hline
    Min   &  0.0399      &  0.0572     &  0.0476       &  0.0501  \\
    Max   &  0.1379      &  0.1883     &  0.1306       &  0.1389  \\
    Avg   &  0.0501      &  0.0951     &  0.0776       &  0.0739  \\
    SD    &  0.0212      &  0.0352     &  0.0211       &  0.0234  \\
\hline
    
    \multicolumn{5}{c}{Training Data (Testing data)}              \\
\hline
    Min   &  0.0155      &  0.0153     &  0.0152       &  0.0144  \\
    Max   &  0.0783      &  0.0332     &  0.0225       &  0.0190  \\
    Avg   &  0.0217      &  0.0201     &  0.0185       &  0.0170  \\
    SD    &  0.0135      &  0.0043     &  0.0021       &  0.0014  \\
\hline
    
    \multicolumn{5}{M{8cm}}{Ratio of the average MAE of testing data to that of training data (majority)}\\
\hline
          &  1.870       &  1.689      &  1.434        &  1.288       \\
\hline
\end{tabular}
\end{table}

The empirical results shown in Tables \ref{understanding route} to \ref{MAE PLM} indicate that, in terms of reducing the vanishing gradient and overfitting, the best version is PLM-LTS-500. We further compared the predictive performance of PLM-LTS-500 with two popular modeling tools: the linear regression model and the conventional backpropagation-based 2LNN model.

The experiment was conducted as follows. First, the linear regression model was used to learn 20 datasets, each of which had 282 training data and 189 testing data. The obtained average MAE for the training and testing data were 0.0142 and 0.0140, respectively. Thus, we set the $\varepsilon$ value of the learning goal of both the conventional 2LNN model and the PLM-LTS-500 model to 0.0282 (i.e., approximately equal to $2 *$ average MAE of the linear regression model). Then, the PLM-LTS-500 model was used to learn the 20 datasets. The final number of hidden nodes for each dataset was recorded and used to set the (fixed) number of hidden nodes for the conventional 2LNN model for each dataset, which is termed the 2LNN\_v model. The (fixed) number of hidden nodes for the other two conventional 2LNN models were set to 13 and 23; the initial weights were set randomly. The number of learning epochs of the backpropagation learning algorithm used to tune the weights of the three conventional 2LNN models was 500.

Table \ref{MAE Compare} shows the MAE results over twenty datasets for the linear regression model, the conventional 2LNN model with 13 hidden nodes (2LNN\_13), the conventional 2LNN model with 23 hidden nodes (2LNN\_23), the 2LNN\_v model, and the PLM-LTS-500 model. Here, for the PLM-LTS-500 model, the training MAE was calculated for the first 272 training data sorted by LTS and the test MAE was calculated for all 189 testing data. Table XII shows that all MAE results for PLM-LT-500 were better than those for the conventional 2LNN models. Thus, PLM yields much better learning performance than backpropagation learning. Furthermore, all MAE results for PLM-LT-500 are better than those for the linear regression model under the condition that the $\varepsilon$ value of the learning goal is approximately equal to twice the average MAE of the linear regression model. Thus, the proposed PLM-LTS-500 model yields performance superior to the linear regression model and the conventional 2LNN model.

\begin{table}
\renewcommand{\arraystretch}{1.3}
\caption{MAE Results for Linear Regression Model, 2LNN\_13, 2LNN\_23, 2LNN\_v and PLM-LTS-500.}
\label{MAE Compare}
\centering
\begin{tabular}{cM{1.3cm}cccM{1cm}}
\hline
          &  Linear Regression  &  2LNN\_13  &  2LNN\_23  &  2LNN\_v  &  PLM-LTS-500\\
\hline
    \multicolumn{6}{c}{Training Data}\\
\hline
    Min   &  0.0134             &  0.0073    &  0.0069    &  0.0073   &  0.0066\\
    Max   &  0.0150             &  0.2056    &  0.2321    &  0.2044   &  0.0090\\
    Avg   &  0.0142             &  0.0820    &  0.0741    &  0.0974   &  0.0079\\
    SD    &  0.0005             &  0.0645    &  0.0667    &  0.0517   &  0.0008\\
\hline
    \multicolumn{6}{c}{Testing Data}\\
\hline
    Min   &  0.0127             &  0.0114    &  0.0107    &  0.0123   &  0.0121\\
    Max   &  0.0155             &  0.1855    &  0.2172    &  0.2181   &  0.0149\\
    Avg   &  0.0140             &  0.0813    &  0.0744    &  0.0980   &  0.0138\\
    SD    &  0.0008             &  0.0569    &  0.0620    &  0.0509   &  0.0008\\
\hline
\end{tabular}
\end{table}

\section{Conclusion}
\noindent In this study we conduct experiments using a copper price forecasting dataset to validate and evaluate the proposed PLM. The experimental results shown in Fig. \ref{Fig:hidden node 10} validate that the proposed PLM indeed implements the pupil learning procedure of Table \ref{plp} with interpretating, picking, understanding, cramming, and organizing. For instance, the interpreting\_LTS module separates all instances into the acquainted instances and the unacquainted instances. The picking\_LTS module picks unacquainted instances one by one such that easy (unacquainted) instances are learned first and similar instances are grouped together. The cramming module enforces memorization when the understanding module cannot learn a picked unacquainted instance. The organizing module comprehends all learned instances for a concise knowledge system. The empirical results show that these modules constitute sequential learning, adaptive learning, perfect learning, and less-overfitted learning. The empirical results also verify the effectiveness of the proposed PLM in handling both vanishing gradients and overfitting.

The empirical results of the second experiment also attest the superiority of the PLM model over the linear regression model and the conventional backpropagation-based 2LNN model. The empirical results indicate that the learning process resulting from the PLM varies based on the hyperparameter arrangements of all modules. One future work is to identify the best hyperparameters for all modules for specific applications when the PLM is adopted.

\section{Acknowledgments}
\noindent This research was supported by the Ministry of Science and Technology, Taiwan, under grants MOST 109-2410-H-004-067-MY2 and 111-2410-H-004-063-MY3.

% ##############################################

\bibliographystyle{IEEEtran}
\bibliography{references}

\begin{thebibliography}{10}
\providecommand{\url}[1]{#1}
\csname url@samestyle\endcsname
\providecommand{\newblock}{\relax}
\providecommand{\bibinfo}[2]{#2}
\providecommand{\BIBentrySTDinterwordspacing}{\spaceskip=0pt\relax}
\providecommand{\BIBentryALTinterwordstretchfactor}{4}
\providecommand{\BIBentryALTinterwordspacing}{\spaceskip=\fontdimen2\font plus
\BIBentryALTinterwordstretchfactor\fontdimen3\font minus
  \fontdimen4\font\relax}
\providecommand{\BIBforeignlanguage}[2]{{%
\expandafter\ifx\csname l@#1\endcsname\relax
\typeout{** WARNING: IEEEtran.bst: No hyphenation pattern has been}%
\typeout{** loaded for the language `#1'. Using the pattern for}%
\typeout{** the default language instead.}%
\else
\language=\csname l@#1\endcsname
\fi
#2}}
\providecommand{\BIBdecl}{\relax}
\BIBdecl

\bibitem{rumelhart2013learning}
D.~Rumelhart, G.~Hinton, and R.~Williams, ``Learning internal representations
  by error propagation,[readings in cognitive science: A perspective from
  psychology and artificial intelligence],'' 2013.

\bibitem{perez2018review}
B.~P{\'e}rez-S{\'a}nchez, O.~Fontenla-Romero, and B.~Guijarro-Berdi{\~n}as, ``A
  review of adaptive online learning for artificial neural networks,''
  \emph{Artificial Intelligence Review}, vol.~49, pp. 281--299, 2018.

\bibitem{ma2003new}
L.~Ma and K.~Khorasani, ``A new strategy for adaptively constructing multilayer
  feedforward neural networks,'' \emph{Neurocomputing}, vol.~51, pp. 361--385,
  2003.

\bibitem{mezard1989learning}
M.~Mezard and J.-P. Nadal, ``Learning in feedforward layered networks: The
  tiling algorithm,'' \emph{Journal of Physics A: Mathematical and General},
  vol.~22, no.~12, p. 2191, 1989.

\bibitem{kusuma1992cascade}
T.~Kusuma and M.~M. Brown, ``Cascade-correlation learning architecture for
  first-break picking and automated trace editing,'' in \emph{SEG Technical
  Program Expanded Abstracts 1992}.\hskip 1em plus 0.5em minus 0.4em\relax
  Society of Exploration Geophysicists, 1992, pp. 1136--1139.

\bibitem{chen1994generating}
Y.~Q. Chen, D.~W. Thomas, and M.~S. Nixon, ``Generating-shrinking algorithm for
  learning arbitrary classification,'' \emph{Neural Networks}, vol.~7, no.~9,
  pp. 1477--1489, 1994.

\bibitem{frean1990upstart}
M.~Frean, ``The upstart algorithm: A method for constructing and training
  feedforward neural networks,'' \emph{Neural computation}, vol.~2, no.~2, pp.
  198--209, 1990.

\bibitem{tsaih1993softening}
R.~Tsaih, ``The softening learning procedure,'' \emph{Mathematical and computer
  modelling}, vol.~18, no.~8, pp. 61--64, 1993.

\bibitem{tsaih1998explanation}
------, ``An explanation of reasoning neural networks,'' \emph{Mathematical and
  Computer Modelling}, vol.~28, no.~2, pp. 37--44, 1998.

\bibitem{tsaih2009resistant}
R.-H. Tsaih and T.-C. Cheng, ``A resistant learning procedure for coping with
  outliers,'' \emph{Annals of Mathematics and Artificial Intelligence},
  vol.~57, pp. 161--180, 2009.

\bibitem{tsai2019cramming}
Y.-H. Tsai, Y.-J. Jheng, and R.-H. Tsaih, ``The cramming, softening and
  integrating learning algorithm with parametric relu activation function for
  binary input/output problems,'' in \emph{2019 International Joint Conference
  on Neural Networks (IJCNN)}.\hskip 1em plus 0.5em minus 0.4em\relax IEEE,
  2019, pp. 1--7.

\bibitem{dietterich1995overfitting}
T.~Dietterich, ``Overfitting and undercomputing in machine learning,''
  \emph{ACM computing surveys (CSUR)}, vol.~27, no.~3, pp. 326--327, 1995.

\bibitem{hawkins2004problem}
D.~M. Hawkins, ``The problem of overfitting,'' \emph{Journal of chemical
  information and computer sciences}, vol.~44, no.~1, pp. 1--12, 2004.

\bibitem{amari1993backpropagation}
S.-i. Amari, ``Backpropagation and stochastic gradient descent method,''
  \emph{Neurocomputing}, vol.~5, no. 4-5, pp. 185--196, 1993.

\bibitem{abadi2016tensorflow}
M.~Abadi, A.~Agarwal, P.~Barham, E.~Brevdo, Z.~Chen, C.~Citro, G.~S. Corrado,
  A.~Davis, J.~Dean, M.~Devin \emph{et~al.}, ``Tensorflow: Large-scale machine
  learning on heterogeneous distributed systems,'' \emph{arXiv preprint
  arXiv:1603.04467}, 2016.

\bibitem{abadi2016tensorflow2}
M.~Abadi, P.~Barham, J.~Chen, Z.~Chen, A.~Davis, J.~Dean, M.~Devin,
  S.~Ghemawat, G.~Irving, M.~Isard \emph{et~al.}, ``Tensorflow: A system for
  large-scale machine learning. 12th $\{$USENIX$\}$ symposium on operating
  systems design and implementation ($\{$OSDI$\}$ 16), 265--283,'' \emph{Google
  Scholar Google Scholar Digital Library Digital Library}, 2016.

\bibitem{paszke2019pytorch}
A.~Paszke, S.~Gross, F.~Massa, A.~Lerer, J.~Bradbury, G.~Chanan, T.~Killeen,
  Z.~Lin, N.~Gimelshein, L.~Antiga \emph{et~al.}, ``Pytorch: An imperative
  style, high-performance deep learning library,'' \emph{Advances in neural
  information processing systems}, vol.~32, 2019.

\bibitem{bengio2013representation}
Y.~Bengio, A.~Courville, and P.~Vincent, ``Representation learning: A review
  and new perspectives,'' \emph{IEEE transactions on pattern analysis and
  machine intelligence}, vol.~35, no.~8, pp. 1798--1828, 2013.

\bibitem{lecun2015deep}
Y.~LeCun, Y.~Bengio, G.~Hinton \emph{et~al.}, ``Deep learning. nature, 521
  (7553), 436-444,'' \emph{Google Scholar Google Scholar Cross Ref Cross Ref},
  p.~25, 2015.

\bibitem{lu2017adaptive}
S.-M. Lu, D.-P. Li, and Y.-J. Liu, ``Adaptive neural network control for
  uncertain time-varying state constrained robotics systems,'' \emph{IEEE
  Transactions on Systems, Man, and Cybernetics: Systems}, vol.~49, no.~12, pp.
  2511--2518, 2017.

\bibitem{ma2019adaptive}
L.~Ma and L.~Liu, ``Adaptive neural network control design for uncertain
  nonstrict feedback nonlinear system with state constraints,'' \emph{IEEE
  Transactions on Systems, Man, and Cybernetics: Systems}, vol.~51, no.~6, pp.
  3678--3686, 2019.

\bibitem{niu2018adaptive}
B.~Niu, H.~Li, Z.~Zhang, J.~Li, T.~Hayat, and F.~E. Alsaadi, ``Adaptive
  neural-network-based dynamic surface control for stochastic interconnected
  nonlinear nonstrict-feedback systems with dead zone,'' \emph{IEEE
  Transactions on Systems, Man, and Cybernetics: Systems}, vol.~49, no.~7, pp.
  1386--1398, 2018.

\bibitem{wang2020multiscale}
X.~Wang, A.~Bao, E.~Lv, and Y.~Cheng, ``Multiscale multipath ensemble
  convolutional neural network,'' \emph{IEEE Transactions on Systems, Man, and
  Cybernetics: Systems}, vol.~51, no.~9, pp. 5918--5928, 2020.

\bibitem{hara2015analysis}
K.~Hara, D.~Saito, and H.~Shouno, ``Analysis of function of rectified linear
  unit used in deep learning,'' in \emph{2015 international joint conference on
  neural networks (IJCNN)}.\hskip 1em plus 0.5em minus 0.4em\relax IEEE, 2015,
  pp. 1--8.

\bibitem{xu2015empirical}
B.~Xu, N.~Wang, T.~Chen, and M.~Li, ``Empirical evaluation of rectified
  activations in convolutional network,'' \emph{arXiv preprint
  arXiv:1505.00853}, 2015.

\bibitem{kingma2014adam}
D.~P. Kingma and J.~Ba, ``Adam: A method for stochastic optimization,''
  \emph{arXiv preprint arXiv:1412.6980}, 2014.

\bibitem{alameer2019forecasting}
Z.~Alameer, M.~A. Elaziz, A.~A. Ewees, H.~Ye, and Z.~Jianhua, ``Forecasting
  copper prices using hybrid adaptive neuro-fuzzy inference system and genetic
  algorithms,'' \emph{Natural Resources Research}, vol.~28, pp. 1385--1401,
  2019.

\bibitem{liu2017forecasting}
C.~Liu, Z.~Hu, Y.~Li, and S.~Liu, ``Forecasting copper prices by decision tree
  learning,'' \emph{Resources Policy}, vol.~52, pp. 427--434, 2017.

\bibitem{orlowski2017volatility}
L.~T. Orlowski, ``Volatility of commodity futures prices and market-implied
  inflation expectations,'' \emph{Journal of International Financial Markets,
  Institutions and Money}, vol.~51, pp. 133--141, 2017.

\bibitem{wang2019spillover}
T.~Wang and C.~Wang, ``The spillover effects of china's industrial growth on
  price changes of base metal,'' \emph{Resources Policy}, vol.~61, pp.
  375--384, 2019.

\bibitem{astudillo2020copper}
G.~Astudillo, R.~Carrasco, C.~Fern{\'a}ndez-Campusano, and M.~Chac{\'o}n,
  ``Copper price prediction using support vector regression technique,''
  \emph{applied sciences}, vol.~10, no.~19, p. 6648, 2020.

\end{thebibliography}

\newpage

 
\vspace{11pt}

\begin{IEEEbiography}[{\includegraphics[width=1in,height=1.25in,clip,keepaspectratio]{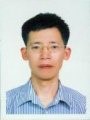}}]{Rua-Huan Tsaih}
received his PhD in operations research in 1991 from the University of California, Berkeley. His research interests include neural networks, decision support systems, operations research, and service innovation. His recent work has been published in Decision Support Systems, Computer Communications, Mathematical and Computer Modelling, Computer \& Operation Research, Communications of the ACM, IT Professional, Expert Systems with Applications, Industrial Management and Data Systems, Malaysian Journal of Library \& Information Science, NTU Management Review, and Review of Securities and Futures Markets.
\end{IEEEbiography}

\begin{IEEEbiography}[{\includegraphics[width=1in,height=1.25in,clip,keepaspectratio]{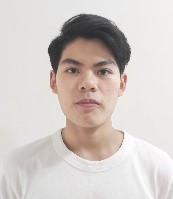}}]{Yu-Hang Chien}
is an undergraduate student at National Chengchi University of Taipei, Taiwan. His research interests include neural networks, computer vision, and distributed systems.
\end{IEEEbiography}

\begin{IEEEbiography}[{\includegraphics[width=1in,height=1.25in,clip,keepaspectratio]{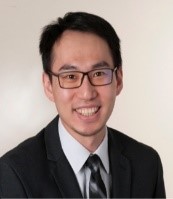}}]{Shih-Yi Chien} is an Associate Professor in the Department of Management Information Systems at National Chengchi University, Taiwan. His current research interests include human-robot interaction, human-automation collaboration, trust in automation, and XAI. His research has appeared in IEEE Transactions on Human- Machine Systems, ACM Transactions on Interactive Intelligent Systems, International Journal of Human Computer Studies, International Journal of Human-Computer Interaction, Electronic Commerce Research and Applications, Journal of Cognitive Engineering and Decision Making, Journal of Intelligent and Robotic Systems, Human Factors, and others.
\end{IEEEbiography}

\vspace{11pt}

\vfill

\end{document}